\documentclass{amia}
\usepackage[T1]{fontenc}
\usepackage{lipsum} 

\begin{document}

\title{Multi-Task Learning for Extracting Menstrual Characteristics \\ from Clinical Notes}

\author{Anna Shopova, BS$^1$, Christoph Lippert, PhD$^1$, Leslee J. Shaw, PhD$^2$, \\Eugenia Alleva, MD, MS$^3$$^,$$^4$$^,$$^5$}

\institutes{
    $^1$Hasso Plattner Institute, Digital Engineering Faculty, University of Potsdam, Potsdam, Germany; $^2$Blavatnik Family Women's Health Research Institute, Icahn School of Medicine at Mount Sinai, New York, USA$^3$Hasso Plattner Institute for Digital Health at Mount Sinai, Icahn School of Medicine at Mount Sinai, New York, USA, $^3$ Department for Artificial Intelligence and Human Health, Icahn School of Medicine at Mount Sinai, New York, USA, $^5$Department of Obstetrics, Gynecology, and Reproductive Science, Icahn School of Medicine at Mount Sinai, New York, USA
}

\maketitle

\section*{Abstract}

\textit{Menstrual health is a critical yet often overlooked aspect of women’s healthcare. Despite its clinical relevance, detailed data on menstrual characteristics is rarely available in structured medical records. To address this gap, we propose a novel Natural Language Processing pipeline to extract key menstrual cycle attributes - dysmenorrhea, regularity, flow volume, and intermenstrual bleeding. Our approach utilizes the GatorTron model with Multi-Task Prompt-based Learning, enhanced by a hybrid retrieval preprocessing step to identify relevant text segments. It outperforms baseline methods, achieving an average F1-score of 90\% across all menstrual characteristics, despite being trained on fewer than 100 annotated clinical notes. The retrieval step consistently improves performance across all approaches, allowing the model to focus on the most relevant segments of lengthy clinical notes. These results show that combining multi-task learning with retrieval improves generalization and performance across menstrual characteristics, advancing automated extraction from clinical notes and supporting women's health research.}

\section{Introduction}
\label{sec:introduction}

Menstrual health is a critical factor in a woman's overall well-being, yet it is often neglected and inadequately documented in routine clinical practice. Recent findings highlight research gaps in women's health, particularly in reproductive and gynecological conditions\cite{nasem2024overview}. Detailed and accurate documentation of menstrual characteristics, such as period cycle regularity, flow volume, and dysmenorrhea severity, is critical to the diagnosis and management of a variety of gynecological conditions \cite{Chodankar22, jain23}. Beyond gynecological disorders, irregular menstrual cycles and heavy bleeding have been associated with an increased risk of cardiometabolic diseases, including hypertension, diabetes, coronary heart disease, and stroke \cite{Keenan23, lo23, wang22, wang24, okoth23}. Heavy menstrual bleeding has been associated with an increased risk of cardiovascular disease, suggesting that it may serve as an early indicator that warrants clinical attention \cite{Dubey24}. Similarly, dysmenorrhea, a common but often overlooked menstrual symptom, has recently been associated with an increased risk of ischemic heart disease and stroke, further highlighting the importance of detailed menstrual documentation in clinical practice \cite{Yeh22, Yeh23, Alleva2024}.

Electronic health records (EHRs) offer an opportunity to investigate menstrual health conditions over time. Although EHRs contain structured information such as diagnostic codes, laboratory values, and procedural codes, these data points often fall short in providing a comprehensive phenotypic characterization of patients \cite{tayefi21}. In addition, menstrual health conditions are often underreported in structured EHR data. This is significant because epidemiological studies of EHRs may have less power if they use only coded diagnoses \cite{Alleva2024}. In contrast, clinical notes contain rich information, including information related to menstrual health captured in routine gynecological visits.

With advances in deep learning, Natural Language Processing (NLP) has emerged as a key method for transforming unstructured clinical narratives into structured data. Various NLP approaches, ranging from supervised fine-tuning (SFT) to zero-shot prompting, have been used to extract information from clinical notes \cite{Khan2024ACS, liu2024, bilal2025, zaghir2024}. However, it is unclear to what extent these approaches are suitable for extracting information on menstrual health, and the optimal method for this task has yet to be determined.

In this work, we explore the effectiveness of different NLP techniques on menstrual characteristics from clinical notes. Our contributions include:
\begin{itemize}
  \item Developing an NLP pipeline specifically designed to extract five clinically relevant menstrual attributes: Presence of Dysmenorrhea, Dysmenorrhea Severity, Regularity, Flow, and Intermenstrual Bleeding.
  \item Comparing the performance of Supervised Fine-Tuning (SFT), In-Context Learning (ICL), and Prompt-Based Learning (PBL) approaches.
  \item Investigating the impact of incorporating a retrieval data preprocessing step, using a hybrid method that combines keyword-based and semantic search, to identify the most relevant text segments from clinical note.
  \item Introducing Multi-Task Prompt-Based Learning (MTPBL) to simultaneously extract multiple menstrual attributes, demonstrating improved generalization and efficiency compared to single-task approaches.
\end{itemize}
By evaluating these methods, we aim to identify the most effective and computationally efficient approach for the extraction of menstrual attributes, ultimately supporting improved clinical decision-making and advancing research in women's health.

\section{Methods}
\label{sec:methods}

\paragraph{Dataset}
\label{subsec:dataset}

Clinical notes from the Electronic Health Records (EHR) of the Mount Sinai Data Warehouse (MSDW) \cite{datawarehouse} were acquired through the Artificial Intelligence Ready Mount Sinai (AIR·MS) platform. Initially, 200 clinical notes associated to a gynecological well-woman visits were randomly extracted. After filtering out notes from patients who were not actively menstruating, the final dataset consisted of 140 clinical notes randomly split in a 65:35 ratio into training (N=91) and test (N=49) set.

Each note was manually annotated by a clinician for dysmenorrhea (yes, no, unknown), dysmenorrhea severity (mild, moderate, severe, unknown), menstrual regularity (regular, irregular, unknown), menstrual flow (scanty, normal, abundant, unknown), and intermenstrual bleeding (yes, no, unknown). These annotations served as the gold standard for the development and evaluation of the proposed NLP pipeline.

Table \ref{tab:dataset} presents the distribution of the labels in the train and test sets. Dysmenorrhea is explicitly mentioned in 52\% of all 140 clinical notes. Among these, the severity is documented in 95\% of the cases labeled "yes", with the remaining 5\% marked as "unknown." Regularity is documented in 78\% of the notes, flow in 64\%, while intermenstrual bleeding is the least frequently documented, mentioned in only 13\% of all clinical notes.
 
\begin{table}[h!]
    \caption{Number of occurrences for each menstrual attribute in the clinical notes for train and test dataset.}
    \label{tab:dataset}
    \centering
    \scalebox{0.9}{
    \begin{tabular}{c @{\hspace{1cm}} c @{\hspace{1cm}} c c}
    \toprule
    Menstrual Attribute & Class & Number of Train Examples & Number of Test Examples\\
    \midrule
    \multirow{3}{4em}{Dysmenorrhea} & yes & 29 & 11 \\
    & no & 21 & 12 \\
    & unknown & 41 & 26 \\
    \cmidrule(lr){1-4}
    \multirow{4}{4em}{Dysmenorrhea severity} & mild & 7 & 7 \\
    & moderate & 11 & 2 \\
    & severe & 10 & 1 \\
    & unknown & 63 & 39 \\
    \cmidrule(lr){1-4}
    \multirow{3}{4em}{Regularity} & regular & 68 & 28 \\
    & irregular & 9 & 4 \\
    & unknown & 14 & 17 \\
    \cmidrule(lr){1-4}
    \multirow{4}{4em}{Flow} & scanty & 3 & 2 \\
    & normal & 46 & 18 \\
    & abundant & 10 & 10 \\
    & unknown & 32 & 19 \\
    \cmidrule(lr){1-4}
    \multirow{3}{4em}{Intermenstrual Bleeding} & yes & 3 & 0 \\
    & no & 11 & 4 \\
    & unknown & 77 & 45 \\
    \bottomrule
    \end{tabular}}
\end{table}

\paragraph{Retrieval}
\label{subsec:retrieval}

\begin{figure}[h!]
  \centering
    \includegraphics[width=1.0\textwidth]{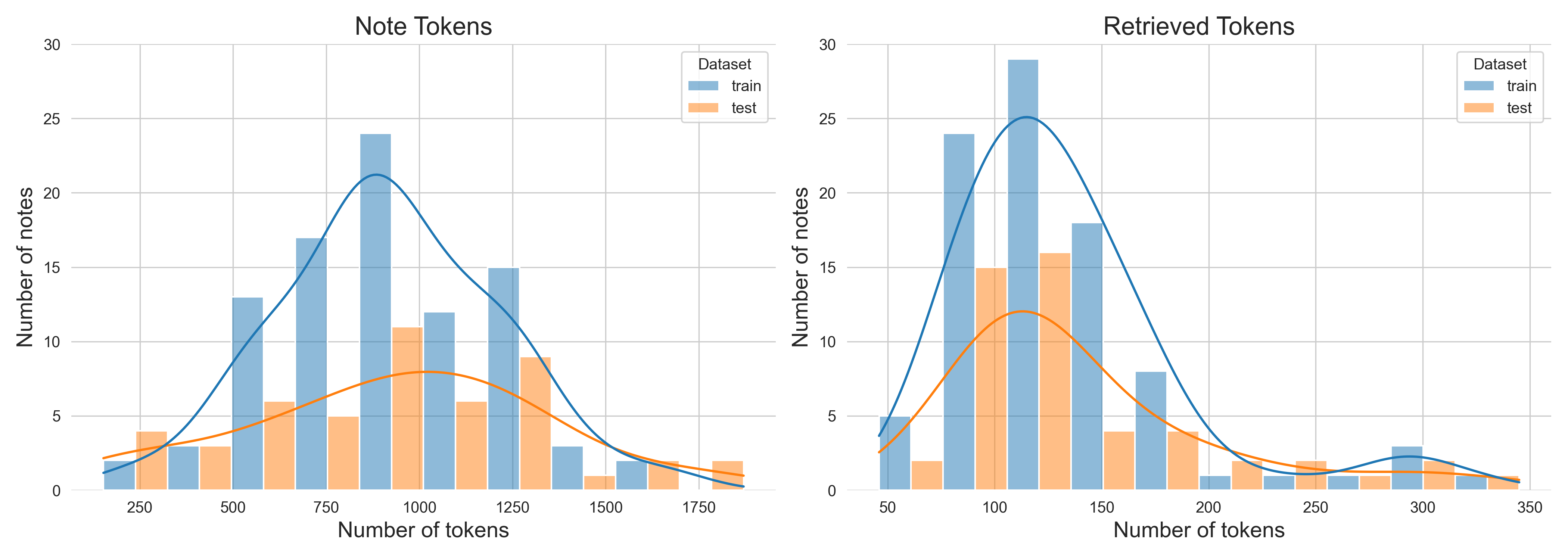}
 \caption{Number of tokens, before and after retrieval of relevant information in both, train and test dataset.}
 \label{fig:note_tokens}
\end{figure}

Clinical notes tend to be comparatively long. As shown in Figure \ref{fig:note_tokens}, most clinical notes in our dataset exceed a token length of 512, which poses a challenge for models with token constraints. Additionally, gynecological clinical notes often contain a large amount of information that is not directly relevant to menstrual health attributes, making it essential to identify and prioritize the most important segments. A naive truncation strategy discards potentially critical information, whereas retrieval allows us to retain and prioritize the most relevant segments for processing. To address this, we implement a hybrid retrieval for preprocessing our datasets, combining BM25 for fast lexical matching and MedEmbed-small-v0.1\cite{balachandran2024medembed} for semantic similarity scoring. 

The clinical notes were segmented using a rule-based approach, splitting text at double spaces to separate sentences, multi-sentence blocks, or bullet points. Each segment was compared to the following predefined retrieval query:
\begin{equation}
    \begin{aligned}
        Query_{retrieval} =  & \quad ''dysmenorrhea, regularity, period\ pattern, menses, flow\ volume,\\
        & \quad bleeding\ pattern, intermenstrual\ bleeding, spotting''
    \end{aligned}
\end{equation}
For each clinical note the top 10 most similar segments were retrieved. This choice of k=10 was based on the assumption that, since we extract four menstrual attributes, it is unlikely that relevant information is spread across more than 10 segments within a single note. Figure \ref{fig:note_tokens} illustrates the distribution of the lengths of the retrieved notes. The majority of retrieved segments contain about 100 tokens, with some containing between 150 and 350 tokens.

\subsection*{Classification Approach}

Each of our five menstrual attributes is treated as a separate classification task, requiring the model to distinguish between predefined categorical labels based on textual descriptions in clinical notes. By structuring the problem in this way, we enable both single-task and multi-task learning approaches, allowing a comparative analysis of their effectiveness in capturing different menstrual characteristics.

\paragraph{Baselines} 
To evaluate the performance of different methods for extracting menstrual attributes, we establish baseline models that serve as reference points for comparison.

\begin{itemize}
    \item \textbf{Supervised Fine-Tuning (SFT)}: Fine-tuning is a widely used approach for text classification tasks. In this work, we used the GatorTron-Base \cite{yang2022gatortronlargeclinicallanguage}, a domain-specific clinical language model. We fine-tuned the model using supervised learning, optimizing for categorical classification of each menstrual attribute. 

    \item \textbf{In-Context Learning (ICL)}: Instead of fine-tuning, ICL allows large language models to perform task-specific classification by providing structured prompts at inference time, without modifying model parameters. For this study, we used Meditron-3 \cite{meditron}, a clinically fine-tuned LLaMA 3.1-based model \cite{grattafiori2024llama3herdmodels}. We constructed a detailed three-shot prompt with examples covering all five menstrual attributes, guiding the model to generate structured predictions within the predefined label space.
\end{itemize}

Both baselines are evaluated with and without retrieval to assess its impact on performance. While retrieval is particularly beneficial for GatorTron-Base due to its 512-token input limit, Meditron-3 has a larger input capacity, allowing it to process more context per pass. However, retrieval remains valuable for prioritizing the most clinically relevant information in lengthy notes, ensuring the models focus on critical menstrual attributes.

\paragraph{Prompt-Based Learning (PBL)}
\label{subsec:prompt-based-learning}

\begin{table}[h!]
    \caption{Task-Specific Prompt Templates and Verbalizers}
    \label{tab:templetes_verbalizers}
    \centering
    \scalebox{0.85}{
    \begin{tabular}{l @{\hspace{1cm}} l l}
    \toprule
    Task & Template & Verbalizer\\
    \midrule
    \multirow{3}{6em}{Dysmenorrhea} & \multirow{3}{20em}{\(Text_i\) dysmenorrhea: \texttt{[MASK]}} & $yes \rightarrow yes, mild, moderate, severe$ \\
    & & $no \rightarrow no, none$ \\
    & & $unknown \rightarrow unknown, unspecified, uncertain$ \\
    \midrule
    \multirow{4}{6em}{Dysmenorrhea severity} & \multirow{3}{20em}{\(Text_i\) dysmenorrhea severity: \texttt{[MASK]}} & $mild \rightarrow mild, light, manageable$ \\
    & & $moderate \rightarrow moderate, medium, average$\\
    & & $severe \rightarrow severe, intense, extreme, painful$\\
    & & $unknown \rightarrow unknown, unspecified, uncertain$ \\
    \midrule
    \multirow{3}{6em}{Regularity} & \multirow{3}{20em}{\(Text_i\) period pattern: \texttt{[MASK]}} & $regular \rightarrow regular, normal$\\
    & & $irregular \rightarrow irregular$\\
    & & $unknown \rightarrow unknown, unspecified, uncertain$\\
    \midrule
    \multirow{4}{6em}{Flow} & \multirow{3}{20em}{\(Text_i\) bleeding pattern: \texttt{[MASK]}} & $scanty \rightarrow scanty, light, spotting$ \\
    & & $normal \rightarrow normal, regular, moderate$\\
    & & $abundant \rightarrow abundant, heavy, profuse$\\
    & & $unknown \rightarrow unknown, unspecified, uncertain$\\
    \midrule
    \multirow{3}{6em}{Intermenstrual Bleeding} & \multirow{3}{20em}{\(Text_i\) intermenstrual bleeding: \texttt{[MASK]}} & $yes \rightarrow yes, present, spotting$ \\
    & & $no \rightarrow no, none, absent$ \\
    & & $unknown \rightarrow unknown, unspecified, uncertain$ \\
    \bottomrule
    \end{tabular}}
\end{table}

PBL leverages pre-trained language models (PLMs) by structuring inputs as prompts that guide the model's predictions. Rather than traditional SFT, PBL uses the masked langauge modeling (MLM) objective in which masked tokens are predicted based on their context in the pre-training phase. By appending prompts including masked tokens, the task is aligned to the pre-training objective, enhancing performance with minimally labeled data. To map the output for the masked token to the label of interest, we use a verbalizer, in which a set of label words/ phrases per label is chosen. The final label is mapped by applying normalization and softmax on the logits for each label word for the model's prediction of the mask token to identify the predicted class.

In this work, we implement PBL with GatorTron-Base\cite{yang2022gatortronlargeclinicallanguage}, a transformer-based PLM trained on clinical texts. We utilize OpenPrompt \cite{ding2021openprompt} as our implementation framework, following the general methodology introduced by Schick and Schütze \cite{schick2021exploiting}. To extract menstrual attributes, we manually design prompt templates that frame the extraction task in a fill-in-the-blank format, allowing the model to predict masked tokens based on contextual information in clinical notes (see Table \ref{tab:templetes_verbalizers}. $Text_i$ represents the input clinical note, and the model predicts the \texttt{[MASK]} token based on its learned representations.

We also apply ClinicalLongformer \cite{yikuan2022} within our PBL approach. This long-sequence transformer model for clinical text employs a sparse attention mechanism, combining local sliding window attention and optional global attention, enabling efficient processing of long documents. Although ClinicalLongformer supports sequences up to 4096 tokens, we set the input length to 2048 tokens, which is sufficient to cover our clinical notes without truncation.

\paragraph{Multi-Task Prompt-Based Learning (MTPBL)}


\begin{figure}[t!]
 \centering   \includegraphics[width=1.0\textwidth]{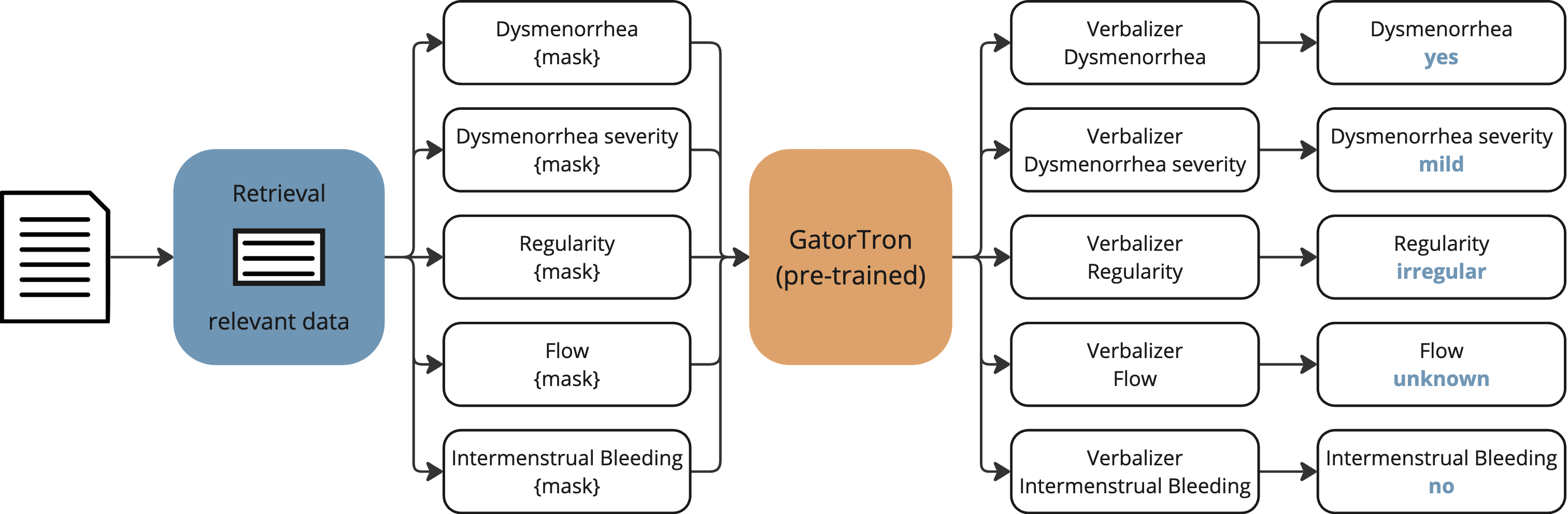}
\caption{NLP Pipeline for extracting and categorizing menstrual characteristics from unstructured clinical notes used in the MTPBL + retrieval approach.}
\label{fig:pipeline}
\end{figure}

Given that each of our five classification tasks relies on the same clinical note corpus, unifying them into a multi-task setup ensures more effective use of shared data and avoids the necessity of training numerous single-task models. To implement this efficient approach, we propose the MTPBL with retrieval, which consists of several sequential steps, as illustrated in Figure~\ref{fig:pipeline}. First, the retrieval method is used to extract relevant text segments from clinical notes, providing input data for all tasks. The task-specific prompt templates are then combined with these retrieved segments to create distinct datasets, each associated with its own data loader. During training, a mini-batch is drawn from each data loader and processed sequentially by a shared PLM (GatorTron-Base), while the order of tasks is shuffled across iterations to mitigate bias towards a specific order of tasks (the data within each mini-batch remains the same). Once the model produces raw logits for each task, a task-specific verbalizer maps these logits to the appropriate label set, and a softmax function then normalizes the output to produce the final predictions. The individual losses computed for each task are summed, averaged, and backpropagated through the shared model in a single optimization step, thereby updating its parameters to improve performance across all tasks simultaneously.

\subsection*{Experimental Setup}
To optimize the hyperparameter of each modeling approach, we employed a 3-fold cross-validation approach on the training dataset, with each fold containing 61 examples for training and 30 for validation. This process was done independently for each modeling approach (SFT, PBL, and MTPBL), both with and without hybrid retrieval (k=10), using macro-averaged F1-score as the primary evaluation metric. After selecting the optimal hyperparameters, we retrained each model on the full training dataset before final evaluation on the test set.

\section{Results}
\label{sec:results}

\begin{table}[h!]
    \caption{Task-specific F1-scores for validation and testing using the different classification approaches presented in this work.}
    \label{tab:results}
    \centering
    \scalebox{0.85}{
    \begin{tabular}{l c c c c c c c c c c}
    \toprule
      &  \multicolumn{2}{c}{\multirow{2}{8em}{\centering Dysmenorrhea}} & \multicolumn{2}{c}{\multirow{2}{8em}{\centering Dysmenorrhea severity}} & \multicolumn{2}{c}{\multirow{2}{8em}{\centering Regularity}} & \multicolumn{2}{c}{\multirow{2}{8em}{\centering Flow}} & \multicolumn{2}{c}{\multirow{2}{8em}{\centering Intermenstrual Bleeding}}\\
      & \\
    \midrule
    \midrule
    \textbf{Validation} & & & & & & & & & &\\
    \midrule
    SFT & 0.732 & & 0.376 & & 0.499 & & 0.354 & & 0.455 &\\
     + retrieval  & 0.805 & \textcolor{ForestGreen}{(+0.073)} & 0.445 & \textcolor{ForestGreen}{(+0.069)} &  0.442 & \textcolor{Gray}{(-0.057)} &  0.449 & \textcolor{ForestGreen}{(+0.095)} &  0.593 & \textcolor{ForestGreen}{(+0.138)}\\
     \midrule
    ICL & 0.705 & & 0.735 & & 0.947 & & 0.281 & & 0.514 &\\
     + retrieval  & 0.910 & \textcolor{ForestGreen}{(+0.205)} & 0.911 & \textcolor{ForestGreen}{(+0.176)}  & 0.861 & \textcolor{Gray}{(-0.086)} & 0.562 & \textcolor{ForestGreen}{(+0.281)} & 0.451 & \textcolor{Gray}{(-0.063)}\\
     \midrule
    PBL & 0.779 &  & 0.804 &  & 0.914 &  & 0.812 &  & 0.613 &\\
     + retrieval  & 0.930 & \textcolor{ForestGreen}{(+0.151)}  & \textbf{0.981} & \textcolor{ForestGreen}{(+0.177)}  & \textbf{0.954} & \textcolor{ForestGreen}{(+0.040)}  & 0.781 & \textcolor{Gray}{(-0.031)} & 0.783 & \textcolor{ForestGreen}{(+0.170)}\\
     \midrule
    MTPBL  & 0.821 &  & \textbf{0.981} &  & 0.817 &  & 0.668 &  & 0.475 &\\
     + retrieval  & \textbf{0.934} & \textcolor{ForestGreen}{(+0.113)} & 0.970 & \textcolor{Gray}{(-0.011)}  & 0.867 & \textcolor{ForestGreen}{(+0.050)}  & \textbf{0.840} & \textcolor{ForestGreen}{(+0.172)}  & \textbf{0.802} & \textcolor{ForestGreen}{(+0.327)}\\
    \midrule
    \midrule
    \textbf{Test} & & & & & & & & & &\\
    \midrule
    SFT & 0.507 & & 0.284 & & 0.406 & & 0.470 & & 0.296 &\\
     + retrieval  & 0.603 & \textcolor{ForestGreen}{(+0.096)} & 0.733 & \textcolor{ForestGreen}{(+0.449)} & 0.570 & \textcolor{ForestGreen}{(+0.164)} & 0.392 & \textcolor{Gray}{(-0.078)} & 0.479 & \textcolor{ForestGreen}{(+0.183)}\\
     \midrule
    ICL & 0.527 & & 0.591 & & 0.796 & & 0.363 & & 0.406 &\\
     + retrieval  & 0.783 & \textcolor{ForestGreen}{(+0.256)} & 0.802 & \textcolor{ForestGreen}{(+0.211)} & 0.823 & \textcolor{ForestGreen}{(+0.027)} & 0.581 & \textcolor{ForestGreen}{(+0.218)} & 0.383 & \textcolor{Gray}{(-0.023)}\\
     \midrule
    PBL & 0.409 & & \textbf{0.914} & & 0.648 & & 0.733 & & 0.823 & \\
     + retrieval  & \textbf{0.888} & \textcolor{ForestGreen}{(+0.479)} & 0.847 & \textcolor{Gray}{(-0.067)} & 0.829 & \textcolor{ForestGreen}{(+0.181)} & 0.750 & \textcolor{Gray}{(+0.017)} & 0.823 & \textcolor{Gray}{($\pm$0.000)}\\
     \midrule
    MTPBL & 0.800 & & 0.836 & & 0.772 & & 0.640 & & 0.823 & \\
     + retrieval & \textbf{0.888} & \textcolor{ForestGreen}{(+0.088)} & 0.891 & \textcolor{ForestGreen}{(+0.055)} & \textbf{0.915} & \textcolor{ForestGreen}{(+0.143)} & \textbf{0.900} & \textcolor{ForestGreen}{(+0.260)} & \textbf{0.923} & \textcolor{ForestGreen}{(+0.100)}\\
    \bottomrule
    \end{tabular}}
\end{table}

\begin{figure}[h!]
 \centering
   \includegraphics[width=1.0\textwidth]{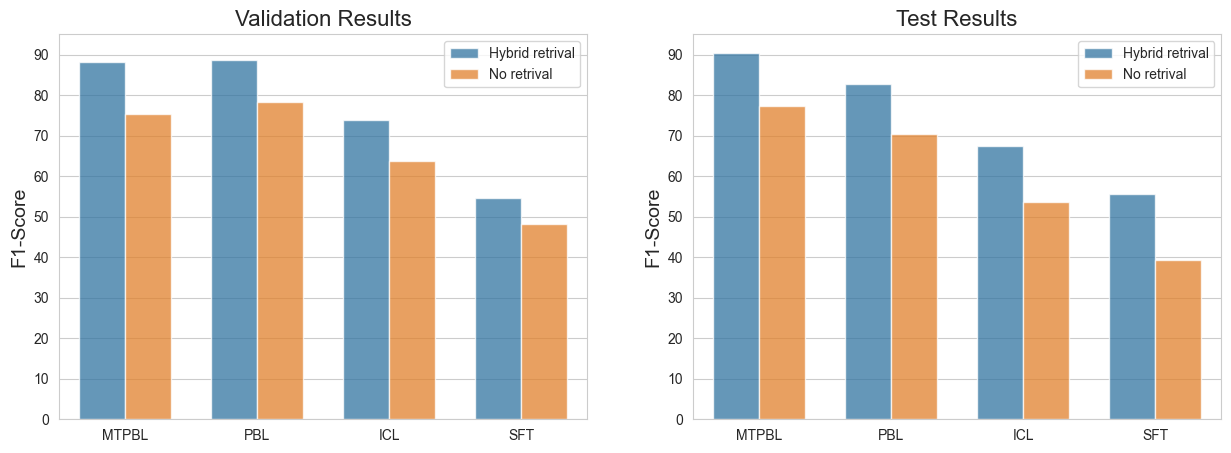}
\caption{Average F1-scores for validation and testing using the different classification approaches presented in this work.}
\label{fig:overall_results}
\end{figure}

Table \ref{tab:results} shows the macro-averaged F1-scores for each method and task, averaged over three cross-validation folds for validation, as well as the test dataset. MTPBL + retrieval secures the highest F1 scores for dysmenorrhea (presence), regularity, flow volume and intermenstrual bleeding in both validation and test. On the test set, this approach gains +0.260 for flow (rising from 0.640 to 0.900) and +0.143 for regularity (0.772 to 0.915) when retrieval is added. A notable exception is dysmenorrhea severity, where PBL without retrieval outperforms other methods on the test data (0.914 vs. 0.847). However, ICL also shows large retrieval-driven gains, especially for dysmenorrhea (+0.205 and +0.256 on test). Overall, MTPBL + retrieval remains the strongest or near strongest performer across all tasks, illustrating how multi-task learning paired with selective text extraction can significantly improve menstrual attribute classification.

Figure \ref{fig:overall_results} shows the macro-averaged F1-scores across all menstrual-attribute tasks for each method, both with and without retrieval. The application of the hybrid retrieval technique, consistently improves performance across all approaches, as shown by the gap between the orange (no retrieval) and blue (hybrid retrieval) bars. On the test set, MTPBL + retrieval achieves the best overall F1-score (0.903), surpassing PBL + retrieval (0.827) and showing that multi-task prompt-based learning generalizes more effectively to unseen data compared to single-task methods.

\begin{table}[h!]
    \caption{F1-scores for validation and testing using ClinicalLongfromer with single-task PBL approach.}
    \label{tab:longformer_results}
    \centering
    \scalebox{0.82}{
    \begin{tabular}{l @{\hspace{1cm}} c @{\hspace{1cm}} c @{\hspace{1cm}} c @{\hspace{1cm}} c @{\hspace{1cm}} c @{\hspace{1cm}} c @{\hspace{1cm}} c}
    \toprule
      &  \multirow{2}{4em}{\centering Dysmenorrhea} & \multirow{2}{4em}{\centering Dysmenorrhea severity} & \multirow{2}{4em}{\centering Regularity} & \multirow{2}{4em}{\centering Flow} & \multirow{2}{4em}{\centering Intermenstrual Bleeding} & \multicolumn{2}{c}{\multirow{2}{8em}{\centering Overall}}\\
      & \\
    \midrule
    \midrule
    \textbf{Validation} & & & & & & &\\
    \midrule
     ClinicalLongformer & 0.599 & 0.456 & \textbf{0.920} & 0.729 & \textbf{0.613} & 0.663 &\\
     + retrieval  & \textbf{0.915} & \textbf{0.925} & 0.883 & \textbf{0.777} & 0.552 & \textbf{0.810} & \textcolor{ForestGreen}{(+0.147)}\\
    \midrule
    \midrule
    \textbf{Test} & & & & & &\\ 
    \midrule
      ClinicalLongformer & 0.231 & 0.308 & 0.378 & 0.380 & \textbf{0.645} & 0.389 &\\
     + retrieval  & \textbf{0.927} & \textbf{0.891} & \textbf{0.794} & \textbf{0.718} & \textbf{0.645} & \textbf{0.795} & \textcolor{ForestGreen}{(+0.406)}\\
    \bottomrule
    \end{tabular}}
\end{table}

Although our primary experiments focused on GatorTron, we also evaluated ClinicalLongformer. In Table \ref{tab:longformer_results} we see that the retrieval step significantly improves ClinicalLongformer’s performance across most attributes, with an overall +0.147 F1 gain in validation and +0.406 in test. The most notable improvement is for dysmenorrhea on the test set, where retrieval increases the F1-score from 0.231 to 0.927 (+0.696), achieving the highest test score among all approaches for this attribute. Without retrieval, ClinicalLongformer shows a drastic performance drop between validation (0.663) and test (0.389), suggesting poor generalization to unseen data. Despite these gains with retrieval, its overall single-task performance remains below that of GatorTron, particularly in regularity, flow, and intermenstrual bleeding. Given these results, we do not consider it for further multi-task training.

\paragraph{Error Analysis}


\begin{table}[hbt!]
    \caption{Examples of incorrect predictions by the MTPBL model with retrieval. Correct answers are marked green while incorrect answers are marked red, followed by the actual correct prediction.}
    \centering
    \scalebox{0.9}{
    \begin{tabularx}{1\textwidth}{>{\hsize=0.1\hsize}X >{\hsize=0.5\hsize}X @{\hspace{1cm}} >{\hsize=0.5\hsize}X}
        \toprule
        Example & Retrieved text & Predictions\\
        \midrule
        1 & \textbf{HMB with last menses}, first time this happened  \textbf{Skips menses frequently}, no menopausal Period Duration (Days) shorter cycles \textbf{Period Pattern Regular} - -  \textbf{Bleeding Pattern normal} - - \textbf{Dysmenorrhea None} - -... & Dysmenorrhea:  \textcolor{ForestGreen}{no} \newline Dysmenorrhea sev.: \textcolor{ForestGreen}{unknown}  \newline Regularity: \textcolor{BrickRed}{regular} \textit{(true: irregular)} \newline Flow: \textcolor{BrickRed}{normal} \textit{(true: abundant)} \newline Intermenstrual Bleeding: \textcolor{ForestGreen}{unknown} \\
        \midrule
        2 & \textbf{Dysmenorrhea} Period Cycle (Days)  Period Duration (Days)  \textbf{Period Pattern  Regular  Regular Regular  Bleeding Pattern} \textbf{\textcolor{BrickRed}{(normal  normal)}}  \textbf{Dysmenorrhea} \textbf{\textcolor{BrickRed}{(None  Moderate)}} GYN: no abnormal vaginal bleeding or discharge... & Dysmenorrhea: \textcolor{BrickRed}{unknown} \textit{(true: yes)} \newline Dysmenorrhea sev.: \textcolor{BrickRed}{unknown} \textit{(true: moderate)} \newline Regularity: \textcolor{ForestGreen}{regular} \newline Flow: \textcolor{ForestGreen}{normal} \newline Intermenstrual Bleeding: \textcolor{ForestGreen}{unknown} \\
        \midrule
        3 & \textbf{Menses are regular} on OTClo.  Menstrual History  Period Cycle (Days): 30  Period Duration (Days):  \textbf{Period Pattern: Regular}  \textbf{Dysmenorrhea: Moderate}  Gynecologic system completely reviewed with age appropriate... & Dysmenorrhea: \textcolor{ForestGreen}{yes} \newline Dysmenorrhea sev.: \textcolor{ForestGreen}{moderate} \newline Regularity: \textcolor{ForestGreen}{regular} \newline Flow: \textcolor{BrickRed}{normal} \textit{(true: unknown)} \newline Intermenstrual Bleeding: \textcolor{ForestGreen}{unknown} \\
        \midrule
        4 & Reason for Visit: Gynecology / Follow-up Visit  Pt has \textbf{monthly menses}, last 6 days, \textbf{moderate, not painful}. Pt denies VD, vaginal itching, dysuria, pelvic pain, breast pain/lumps/ND.  married, is sexually active, denies DV, depression... & Dysmenorrhea: \textcolor{BrickRed}{unknown} \textit{(true: no)} \newline Dysmenorrhea sev.: \textcolor{ForestGreen}{unknown} \newline Regularity: \textcolor{ForestGreen}{regular} \newline Flow: \textcolor{ForestGreen}{abundant} \newline Intermenstrual Bleeding: \textcolor{ForestGreen}{unknown} \\
        \bottomrule
    \end{tabularx}}
    \label{tab:error_examples}
\end{table}

Table \ref{tab:error_examples} presents examples of incorrect predictions made by the MTPBL model with retrieval, highlighting different types of errors observed in the extracted menstrual attributes. In \textbf{Example 1}, inconsistencies occur when conflicting mentions exist within the same note, leading to errors in predicting regularity and flow. \textbf{Example 2} highlights an issue with sentence segmentation, where incorrect text splitting causes missing information in retrieval, resulting in misclassification of dysmenorrhea. Specifically, due to the misinterpretation of double spaces, key details such as "Dysmenorrhea: None Moderate" and "Bleeding Pattern: normal normal" were not properly extracted (highlighted in red), causing the model to miss the necessary context for correct predictions. \textbf{Example 3} shows a case where all necessary information is present, yet the model still predicts "normal" for flow despite no explicit mention in the text. \textbf{Example 4} demonstrates a challenge with the variability of clinical note structure. When menstrual details are documented in a narrative format rather than structured form entries, the model struggles to identify the correct menstrual attributes. Specifically, in associating "not painful" with the absence of dysmenorrhea. These examples illustrate how retrieval effectiveness and text structure impact model performance across different menstrual characteristics.

\subsection{Discussion}
\label{sec:discussion}
Our work explored multiple NLP approaches for extracting menstrual characteristics from clinical notes, demonstrating that prompt-based techniques, particularly Multi-Task Prompt-Based Learning (MTPBL) with retrieval step, achieved the best overall performance. The results highlight key challenges and opportunities in menstrual characteristic extraction, particularly in generalization across tasks, the benefits of retrieval techniques, and the impact of clinical note variability.

Our results show that MTPBL generalizes better than single-task approaches, achieving the highest F1-score across all tasks. Unlike fine-tuning, which is sensitive to small dataset sizes and requires domain-adaptive pre-training, and ICL, which relies heavily on model-internal knowledge, MTPBL effectively utilizes shared task representations. This suggests that while multi-task approaches improve efficiency and robustness, further optimization in task interaction and loss balancing could further enhance performance \cite{chen2024}.

The hybrid retrieval method significantly improved classification across all methods, including fine-tuning and in-context learning, despite their different reliance on model pre-training. Even for ICL, which can process long clinical notes without truncation, retrieval still contributed to better performance by prioritizing clinically relevant information. Similarly, our evaluation of ClinicalLongformer further confirmed the importance our hybrid retrieval preprocessing step. This suggests that retrieval is not only beneficial for overcoming token length constraints but also crucial for structuring and filtering relevant text segments from lengthy clinical notes.

The error analysis showed key challenges in clinical note segmentation. While we initially assumed that a simple rule-based approach, using double spaces as delimiters, would be sufficient on our data, we found that this method did not always work reliably, leading to retrieval errors and incorrect predictions. Given that clinical notes can vary widely in structure, from templated formats to unstructured narrative descriptions, more robust segmentation strategies are needed to ensure consistent extraction \cite{xu2024automaticsentence}.

Another critical observation is the underdocumentation of menstrual characteristics in clinical notes. Despite their clinical significance, certain attributes, such as intermenstrual bleeding, which was explicitly mentioned in only 13\% of our data, or dysmenorrhea, documented in 52\%, are not consistently recorded. This underscores the need for increased awareness among clinicians regarding the importance of documenting menstrual attributes, as their absence can lead to gaps in patient care and limit the ability of NLP models to extract accurate information to support clinical decision-making.

\subsection{Conclusion}
\label{sec:conclusion}

We demonstrate that prompt-based NLP methods, in particular Multi-Task Prompt-Based Learning combined with retrieval-based preprocessing, provide an effective and resource-efficient approach for extracting key menstrual characteristics from clinical notes. While retrieval significantly improves classification performance, challenges remain in addressing variability in clinical notes, segmentation issues, and the manual effort required for prompt templates, verbalizers and retrieval query. Furthermore, the underdocumentation of menstrual characteristics highlights the need for more structured clinical documentation to improve both patient care and NLP-based extraction methods. Future work should focus on reducing manual effort through automated prompt generation and adaptive retrieval, expanding the scope of extracted attributes, and optimizing multi-task learning for greater efficiency and scalability. These improvements will be essential for integrating NLP methods into clinical workflows, improving menstrual health research, and addressing critical gaps in women's health. In addition, validating these methods on larger, multi-institutional datasets and refining adaptive strategies for multi-task learning will be critical to ensuring their robustness and broader applicability.

\makeatletter
\renewcommand{\@biblabel}[1]{\hfill #1.}
\makeatother


\bibliographystyle{vancouver}
\bibliography{amia}

\end{document}